\documentclass[letterpaper, 10 pt, conference]{ieeeconf}

\IEEEoverridecommandlockouts
\usepackage{amsmath}
\usepackage{amssymb}
\usepackage{graphicx}
\usepackage{booktabs}
\usepackage{capt-of}
\usepackage{placeins}
\usepackage{url}
\usepackage{xcolor}

\newcommand{\methodname}{3PoinTr}
\definecolor{myorange}{RGB}{230,120,20}

\title{\LARGE \bf
\methodname: 
From Human Videos to Robot Policies\\  with 3D Point-Track Plans
}

\author{Adam Hung, Bardienus P. Duisterhof, Jeffrey Ichnowski\\
Carnegie Mellon University\\
\\
\textbf{Project page: \textcolor{myorange}{\url{https://adamhung60.github.io/3PoinTr/}}}}

\IEEEaftertitletext{%
\begin{minipage}{\textwidth}
\centering
\vspace{-10pt}
\includegraphics[width=\textwidth]{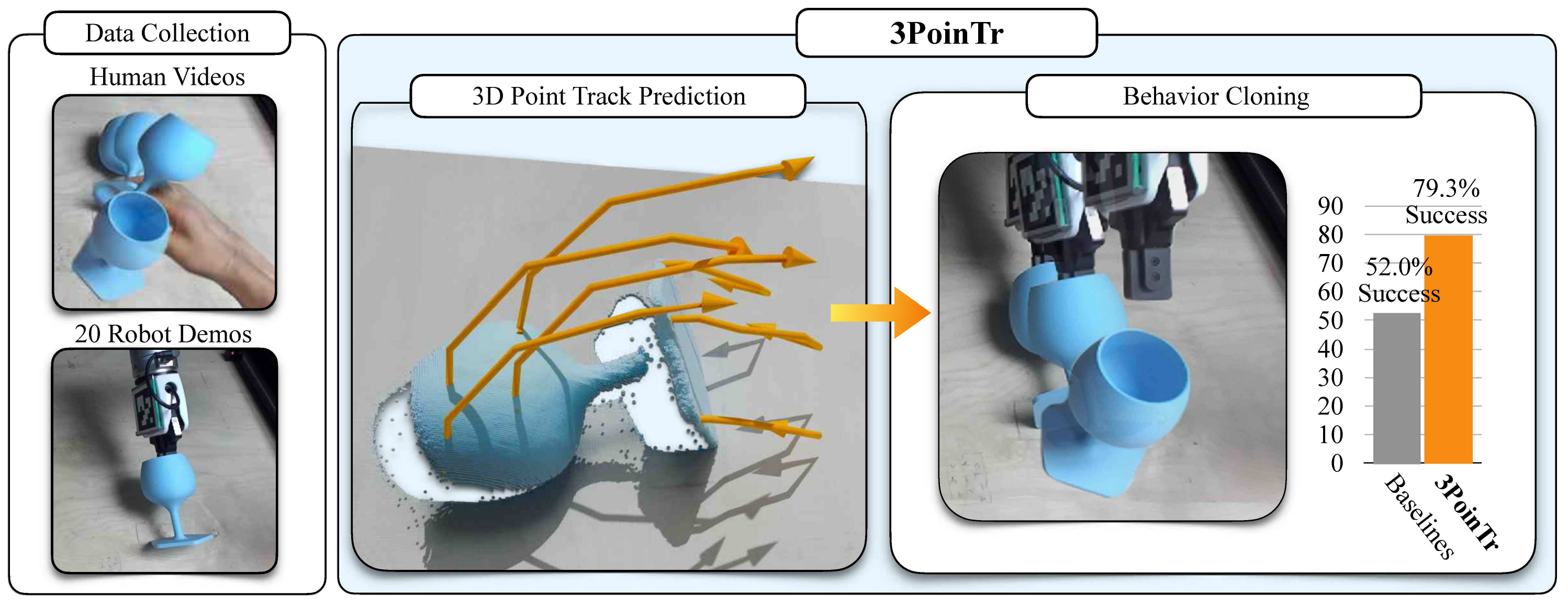}
\captionof{figure}{
  We propose \textbf{\methodname{}}, a method that learns from human videos by treating 3D point tracks as global plans for robot policies.
  From an initial observation, \methodname{} predicts full-task point tracks that provide a global 3D plan to a closed-loop policy.
  A Perceiver-IO encoder compresses this plan into a few plan tokens, which guide the policy throughout execution.
  We find this approach transfers exceptionally well from human demonstrations to robot execution.
  Across eight simulation and real-world tasks with 20 robot demonstrations per task, \methodname{} achieves an average success rate of 79.3\%, compared with 52.0\% for the strongest behavior-cloning or learning-from-video baseline.
  \vspace{10pt}
}
\label{fig:teaser}
\end{minipage}%
}

\begin{document}

\bstctlcite{IEEE:BSTcontrol}

\setcounter{dbltopnumber}{1}

\maketitle
\begin{table*}[t]
\centering
\caption{\textbf{Point-track methods for learning robot control from human
videos.} \checkmark/\(\times\) indicate presence/absence of each property.}
\label{tab:related_work_comparison}
\vspace{-0.5em}
\begingroup
\setlength{\tabcolsep}{5.2pt}
\renewcommand{\arraystretch}{1.12}
\newcommand{\rwyes}{\ensuremath{\checkmark}}
\newcommand{\rwno}{\ensuremath{\times}}
\footnotesize
\begin{tabular}{@{}lcccccc@{}}
\toprule
Method &
\shortstack{3D\\tracks} &
\shortstack{Scene-dense\\tracks} &
\shortstack{Excludes\\embodiment points} &
\shortstack{No object\\segmentation} &
\shortstack{Learned\\policy} &
\shortstack{Tracks predicted\\once} \\
\midrule
\multicolumn{7}{@{}l}{\emph{Using embodiment tracks as robot actions}} \\
Point Policy~\cite{haldar2025point}
  & \rwyes & \rwno  & \rwno  & \rwyes & \rwyes & \rwno \\
Motion Track Policy~\cite{ren_motion_2025}
  & \rwno  & \rwno  & \rwno  & \rwno  & \rwyes & \rwno \\
\midrule
\multicolumn{7}{@{}l}{\emph{Geometrically deriving robot actions from scene motion}} \\
Track2Act~\cite{bharadhwaj_track2act_2024}
  & \rwno  & \rwyes & \rwno  & \rwyes & \rwyes & \rwyes \\
General Flow~\cite{yuan2024general}
  & \rwyes & \rwno  & \rwyes & \rwno  & \rwno  & \rwno \\
3DFlowAction~\cite{zhi_3dflowaction_2025}
  & \rwyes & \rwno  & \rwyes & \rwno  & \rwno  & \rwno \\
TraceGen~\cite{lee_tracegen_2026}
  & \rwyes & \rwyes & \rwno  & \rwyes & \rwno  & \rwyes \\
\midrule
\multicolumn{7}{@{}l}{\emph{Conditioning a learned robot policy on predicted motion}} \\
ATM~\cite{wen_any-point_2024}
  & \rwno  & \rwno  & \rwno  & \rwyes & \rwyes & \rwno \\
AMPLIFY~\cite{amplify}
  & \rwno  & \rwyes & \rwno  & \rwyes & \rwyes & \rwno \\
\(\mu_0\)~\cite{lee_mu0_2026}
  & \rwyes & \rwno  & \rwno  & \rwno  & \rwyes & \rwno \\
Im2Flow2Act~\cite{xu_flow_2024}
  & \rwno  & \rwno  & \rwyes & \rwno  & \rwyes & \rwyes \\
\textbf{\methodname{} (ours)}
  & \rwyes & \rwyes & \rwyes & \rwyes & \rwyes & \rwyes \\
\bottomrule
\end{tabular}
\par\smallskip
\parbox{0.96\textwidth}{
\vspace{10 pt}
\footnotesize
\emph{Scene-dense tracks} cover the scene rather than only an object or selected semantic points.
\emph{No object segmentation} means no object masks or bounding boxes are
required during data processing or execution.
\emph{Learned policy} means a learned model predicts actions, rather than
converting scene motion to actions geometrically.
\emph{Tracks predicted once} means one initial prediction is retained during
execution, as opposed to re-predicting tracks at each time step.}
\endgroup
\vspace{-1em}
\end{table*}

\thispagestyle{empty}
\pagestyle{empty}

\begin{abstract}

Human videos could reduce the reliance on expensive robot demonstrations, but extracting task-relevant features remains challenging.
We introduce \methodname{}, a method that provides robot policies with a
global 3D point-track plan predicted from human videos.
Given an initial point cloud, \methodname{} predicts dense 3D point tracks
spanning the full task and compresses them into a small
set of tokens that remain fixed throughout execution.
At each policy query, the policy combines these plan tokens with the current observation to predict robot actions.
This formulation enables the policy to extract useful motion cues from human videos.
We find this formulation transfers exceptionally well, even when human demonstration motions are infeasible for robots.
In controlled simulation and real-world evaluations, \methodname{} achieves an average success rate \textbf{25.0} percentage points higher than
the strongest behavior-cloning and video-pretraining baselines on real-world
tasks, and an improvement of \textbf{29.6} percentage points in simulation. 
Ablations establish the benefits of our method, including
predicting non-embodiment 3D tracks once as a global plan, Perceiver-IO compression, and a residual cross-attention mechanism.
Our point-track prediction network also outperforms a strong 3D-flow baseline by
retaining supervision for temporarily occluded points.

\end{abstract}

\section{Introduction}

A typical approach to teaching a robot a new task is to collect teleoperation demonstrations and train a policy via behavior cloning. 
However, collecting sufficient demonstration data for each task is expensive~\cite{intelligence2025pi,
oneill_open_2025}.
Human demonstration videos are much more plentiful and easy to collect, and encode the precise scene motion that defines tasks. 
However, humans naturally perform tasks differently from robots, often employing different strategies, trajectories, and grasps.
Therefore, a key challenge is to transfer information from human
demonstrations to robot control without requiring human and robot motion to precisely match.

Point tracks are a promising intermediate representation between human and
robot motion because they describe task-relevant motion explicitly
and can be extracted autonomously from videos. Recent methods learn point tracks from human videos and use them either to directly solve for robot actions or to condition learned policies.
However, many such methods make limiting assumptions about how the predicted motion should translate into robot actions.
Additionally, we find that closed-loop performance is sensitive to
several critical design choices: whether tracks are predicted once as a global plan or re-predicted at each policy query, whether they are represented in 2D or 3D, and how they are compressed and integrated into the policy architecture.
Table~\ref{tab:related_work_comparison} compares prior methods along related axes.

We build on this line of work by treating predicted scene motion as a
\emph{global point-track plan}
rather than an action target. In
\textbf{\methodname}, tracks learned from human videos provide task-critical
motion cues, while teleoperation data teaches the closed-loop policy to predict robot-specific actions
that realize those cues.
\methodname\ first consumes the initial point cloud and predicts dense non-embodiment 3D point tracks that span the
full task execution. A Perceiver-IO encoder~\cite{jaegle_perceiver_2022} compresses these tracks into a small set
of plan tokens, which are cached and held fixed throughout execution. At each
policy query, the closed-loop policy conditions on the plan tokens together with the latest point cloud and robot state to predict the next action.

This paper contributes:
\begin{itemize}

\item A method for transferring motion cues from human videos to robot
control without requiring human and robot trajectories to match.

\item Ablations isolating the design choices that differentiate \methodname{} from prior work:
predicting point tracks once as a global plan,
3D tracks rather than 2D,
predicting non-embodiment point tracks,
Perceiver-IO compression into compact plan tokens,
and residual track-token cross-attention in the action head.

\item A 3D point-track predictor that outperforms a strong 3D-flow
baseline~\cite{yuan2024general} on real-world videos by retaining supervision for
temporarily occluded points.

\end{itemize}

\begin{figure*}[t]
  \centering
  \includegraphics[width=1.0\linewidth]{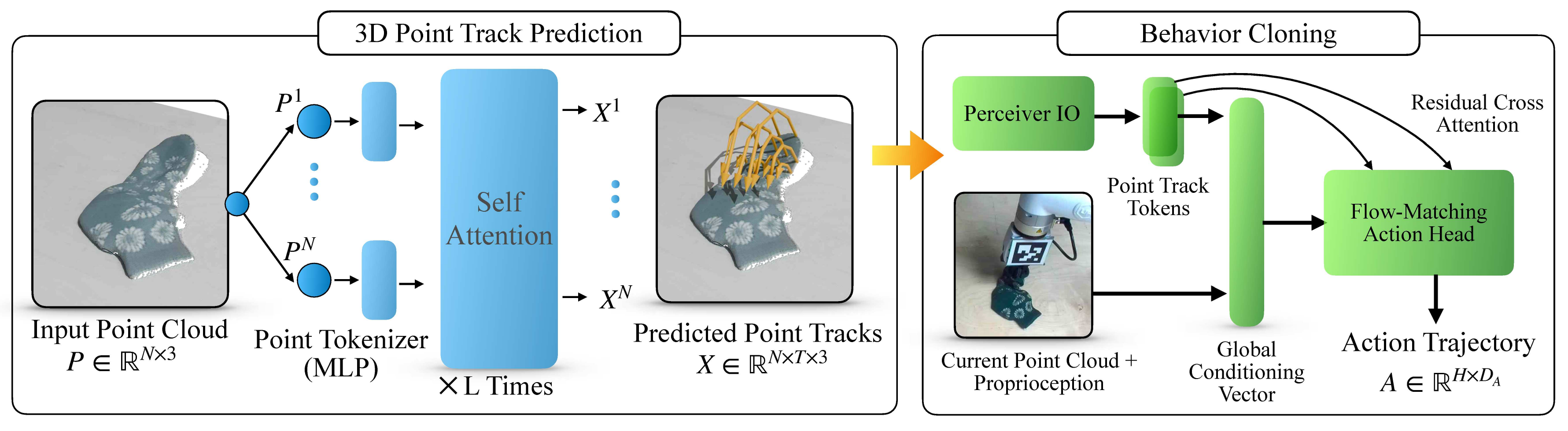}
  \caption{Diagram of the \methodname\ network architecture.
  Given an initial point cloud, a transformer predicts dense 3D point tracks
  spanning the full task execution.
  A Perceiver-IO point-track encoder compresses these tracks into a small set of
  learned plan tokens, which remain fixed throughout the episode.
  At each action step, the policy conditions on the plan tokens together with the
  current point cloud and robot state, enabling closed-loop action prediction.
  Beyond global feature conditioning, residual track-token cross-attention in every
  U-Net block gives the action head a direct path to the predicted object
  motion, encouraging a simpler mapping from point tracks to robot actions.
  }
  \label{fig:arch}
\end{figure*}

\section{Related Work}
\label{sec:related_work}
\methodname{} builds on prior work on behavior cloning and learning from videos.

\paragraph{Behavior cloning}
Behavior cloning succeeds across a wide variety of real-world
tasks~\cite{chi_diffusion_2024,zhao_learning_2023}, but requires substantial
teleoperated data to achieve robustness. Structured representations such as
point clouds~\cite{ze_3d_2024,zhu_point_2024} or
keypoints~\cite{vecerik_robotap_2023} can improve sample efficiency, but the
policy must still learn to map observations to actions from limited robot
demonstrations.

\paragraph{Learning from videos}
Actionless videos (videos without robot action labels) are easy to collect and encode useful task structure. Prior
work uses them for visual pretraining~\cite{nair_r3m_2022}, reward
learning~\cite{zakka_xirl_2021}, affordance
extraction~\cite{bahl_affordances_2023,chen_vidbot_2025}, and conditioning
robot policies on predicted or latent video
dynamics~\cite{liang_video_2025,pai_mimic-video_2025}. 
Similarly, \methodname{} transfers an explicit motion
representation from human videos to robot control.

\paragraph{Learning from videos with point tracks}
Point tracks provide a natural bridge between actionless videos and robot
policies, encoding scene motion in an interpretable, spatially grounded form.
Table~\ref{tab:related_work_comparison} summarizes the key properties of existing point-track-based methods. We group these methods
into three broad categories according to how they use tracks for robot control:
using embodiment tracks as robot actions, geometrically deriving robot actions from scene
motion, and conditioning a learned robot policy.
Related 3D and 2D point methods fall outside this scope because they
either use tracks only as auxiliary representations or do not learn robot
control from human videos~\cite{niu_arm4r_2025,cho_egoavflow_2026,
noh_3dfdp_2025,wang_lamp_2026,guo_flowdreamer_2025,li_egowam_2026,
dharmarajan_dream2flow_2025,li_novaflow_2025,ko_avdc_2024,chen_ecflow_2025,
ranasinghe_fofpred_2026,guan_point_2026,cao_tract_2026,chen_flowwam_2026}.

\emph{Using embodiment tracks as robot actions.}
These methods represent actions as tracks of human or robot
end-effector keypoints and convert predicted tracks directly into robot motion.
Point Policy~\cite{haldar2025point} retargets human points to robot points with a fixed geometric mapping, then trains a policy to predict these retargeted points. 
While this enables learning from only human demonstrations, it assumes that replaying human motion on the robot is both feasible and sufficient to solve the task.
Motion Track Policy~\cite{ren_motion_2025} instead co-trains on human and robot demonstrations using 2D manipulator-keypoint tracks as a shared action representation, which also assumes that humans and robots perform similar motions in similar states.

\emph{Geometrically deriving robot actions from scene motion.}
These methods convert predicted scene motion into robot actions through explicit
geometric procedures. Doing so requires a separately specified grasp or contact
strategy, supplied through heuristics, manual initialization, scripted
behaviors, or external grasp-prediction modules%
~\cite{bharadhwaj_track2act_2024,yuan2024general,yin_3dmf_2025,
zhi_3dflowaction_2025,lee_tracegen_2026}.
The robot then executes the motion implied by the tracks from this selected
contact configuration. Consequently, these methods cannot learn how to make contact with scenes, only how to follow a predicted motion under fixed contact.

\emph{Conditioning a learned robot policy on predicted motion.}
Rather than directly converting tracks into actions, these methods condition a robot policy on predicted motion, allowing it to learn robot-specific behavior from robot data.
ATM~\cite{wen_any-point_2024} conditions policies on point tracks, whereas AMPLIFY~\cite{amplify}
conditions policies on compressed motion tokens derived from point tracks.
\(\mu_0\)~\cite{lee_mu0_2026} similarly conditions policies on intermediate
features from a 3D point track predictor, but relies on a VLM and semantically
selected keypoints.

ATM, AMPLIFY, and \(\mu_0\) all re-predict scene motion from the current observation at each action prediction step.
This induces a distribution gap because intermediate robot states during execution
can fall outside the human-video training distribution.
Additionally, any temporarily occluded points disappear from the conditioning. 
In contrast, \methodname{} predicts full-task tracks once from the initial
scene and retains their compressed representation as a source of guidance
throughout closed-loop execution.
In our ablations (Table~\ref{tab:ablations}), we find this choice to be crucial for real-world performance.

Im2Flow2Act~\cite{xu_flow_2024} is another related method: it predicts
object flow once and conditions a closed-loop inverse dynamics model to follow the predicted flow. However, Im2Flow2Act operates in 2D, cannot predict scene-dense point tracks, and trains its policy entirely
from simulated robot play data generated with predefined action primitives.
Moreover, Im2Flow2Act treats the predicted object flow as the execution target,
which assumes that object motion from human videos matches how the robot should move the object.

In summary, while recent work has established point tracks as a powerful intermediate representation for learning from videos, \methodname{} introduces a policy-conditioning recipe that does not assume human and robot motion will match, does not require object segmentations or predefined grasps, and
outperforms strong video-pretrained baselines in our experiments.

\section{Method}
\label{sec:method}

\methodname\ (Figure~\ref{fig:arch}) consists of two stages: (1) learning a
model that predicts dense future 3D point trajectories from a single scene
observation, and (2) learning a closed-loop policy conditioned on these
predicted point tracks.

\begin{figure*}[t]
    \centering
    \includegraphics[width=1.0\linewidth]{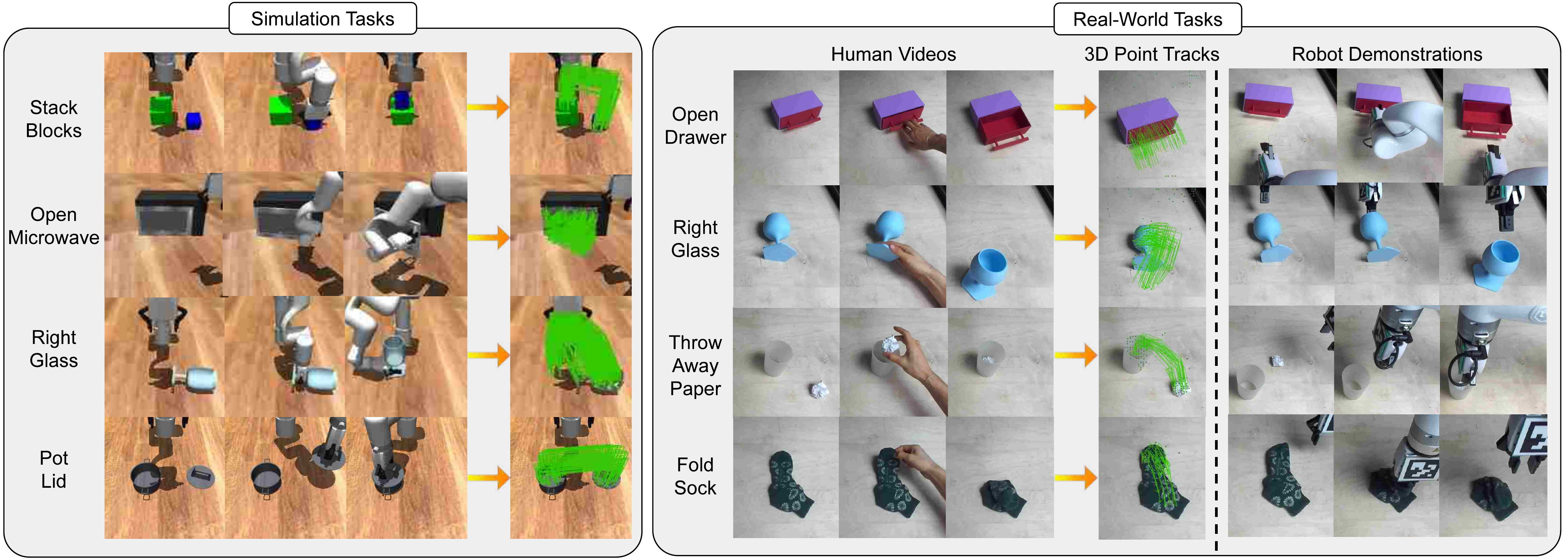}
    \caption{
    Simulation and real-world tasks used for evaluation.
    Simulation tasks use procedurally generated robot trajectories for both video
    pretraining and policy learning; real-world tasks use
    human videos for video pretraining and robot teleoperation data for policy learning.
    Green tracks show 2D projections of the 3D point-track prediction targets (only tracks for moving points are visualized).
    }
    \label{fig:tasks}
\end{figure*}

\subsection{Problem Formulation}

\paragraph{Stage 1: 3D point-track prediction.}
First, given an initial point cloud $P_0 \in \mathbb{R}^{N \times 3}$ of $N$
visible scene points expressed in the camera coordinate frame with all
embodiment points removed, the goal is to predict 3D point tracks: the future 3D
positions of each initial point over a fixed horizon of $T$ timesteps. Formally,
we learn $f_\theta : P_0 \mapsto X$, where
$X = \{X_t^i \mid i \in [0,N), t\in [0,T)\} \in \mathbb{R}^{N \times T \times 3}$, with each
$X^i\in\mathbb{R}^{T\times3}$ corresponding to a prediction for the $i$-th
point in $P_0$ over $T$ time steps.

\paragraph{Stage 2: Track-conditioned policy learning.}

Given the predicted 3D point tracks $X$ together with a current observation
$o_t = (P_t, q_t)$, where $P_t \in \mathbb{R}^{N \times 3}$ is the visible
point cloud at timestep $t$ (including robot points) and $q_t$ is the
robot configuration, we train a closed-loop behavior-cloning policy that predicts a short
chunk of $H$ actions. Formally, we learn
$\pi_\phi : (X, o_t) \mapsto A$, where $A \in \mathbb{R}^{H \times D_A}$ is a chunk of $H$ actions, each with $D_A$ dimensions. The point tracks $X$ are produced
once from the initial observation $P_0$ and reused throughout the episode, while
$o_t$ provides the closed-loop signal: after executing the first
$H_{\mathrm{exec}} \!\le\! H$ predicted actions, the policy is queried again on the
updated $o_t$.

\subsection{Network architecture and training}
\label{sec:arch}

\paragraph{3D point-track prediction.}

We embed visible non-embodiment points with an MLP and pass the resulting tokens
through a transformer with standard self-attention.
A linear prediction head then outputs each point's 3D trajectory over horizon
$T$, yielding point tracks $X \in \mathbb{R}^{N \times T \times 3}$.

During training, we subsample a fixed number of random input points from the initial
point cloud, and apply per-sample unit-sphere normalization.
We assume access to visibility values $m^i_t \in \{0,1\}$, indicating whether
point $i$ is visible at timestep $t$.
We mask losses by $m^i_t$, and use an L1 loss on 3D position error for all
visible point-timestep pairs.

\paragraph{Track-conditioned policy learning.}

During policy training, we freeze the 3D point-track predictor.
Given the first frame of a task, we generate point-track predictions, and then
extract compact yet informative point-track tokens with a \textit{Perceiver-IO}
architecture~\cite{jaegle_perceiver_2022}: each predicted point trajectory
$X^i$ is first embedded into a 3D point-track token $F^i \in \mathbb{R}^{D_F}$,
giving $F \in \mathbb{R}^{N \times D_F}$, and a small set of $M$ learnable
query tokens $Q \in \mathbb{R}^{M \times D_F}$ attends to $F$ via
cross-attention.

Because the policy is closed-loop, we condition it on the current point
cloud $P_t$ (including robot points) and current robot configuration $q_t$ in addition to the persistent track
tokens.
We use a PointNet-style encoder following DP3~\cite{ze_3d_2024} to encode $P_t$,
and a small MLP to encode $q_t$.
The point-track, point-cloud, and configuration features are concatenated and
projected into a global conditioning vector that drives a flow-matching~\cite{lipman2022flow} action head 
built on
the conditional 1D U-Net
architecture of Chi et al.~\cite{chi_diffusion_2024}.

To bias the action head to utilize the predicted object motion rather than 
relying on only the raw
point-cloud observation, we add \textit{residual point-track-token
cross-attention} adapters throughout the U-Net in addition to the 
standard FiLM-style~\cite{perez_film_2017} global conditioning vector.
At each residual block in the down path, middle blocks, and up path, 
the point-track tokens are
linearly projected to the current U-Net channel width, and then the
U-Net features query the point-track tokens. 
The resulting cross-attention output is added back to the U-Net features.
The cross-attention output is zero-initialized, so the U-Net starts
with the same function as the globally conditioned model.
The action head generates action chunks composed of end-effector
position, orientation, and gripper position.
Actions are expressed in the camera coordinate frame.
We train one policy for all simulation tasks
and one policy for all real-world tasks.

\section{Experiments}

\begin{table*}[t]
    \centering
    \setlength{\tabcolsep}{2.6pt}
    \caption{Point-track Average Displacement Error (ADE) and ADE of the
    5\,\% of points that move the most (5\% ADE) for \methodname{} and General
    Flow~\cite{yuan2024general}; lower is better and units are millimeters. 
    On real-world tasks, \methodname{} lowers average ADE by \textbf{28\,\%} and
    5\% ADE by \textbf{44\,\%}, outperforming General Flow on every task and
    metric by retaining supervision for temporarily occluded points. In
    simulation, where there is no occlusion, the methods are competitive.}
    \label{tab:flow_full}
    \begin{tabular}{@{} l rr@{\hspace{0.5em}}rr @{\hspace{1.3em}} l rr@{\hspace{0.5em}}rr @{}}
    \toprule
    \multicolumn{5}{c}{\textbf{Simulation Tasks}} &\multicolumn{5}{c}{\textbf{Real-World Tasks}} \\
    \cmidrule(lr){1-5}
    \cmidrule(l){6-10}
    \textbf{Task} & \multicolumn{2}{c}{\textbf{General Flow}} & \multicolumn{2}{c}{\textbf{\methodname}} & \textbf{Task} & \multicolumn{2}{c}{\textbf{General Flow}} & \multicolumn{2}{c}{\textbf{\methodname}} \\
    \cmidrule(lr){2-3}
    \cmidrule(lr){4-5}
    \cmidrule(lr){7-8}
    \cmidrule(l){9-10}
    & ADE$\downarrow$ & 5\% ADE$\downarrow$ & ADE$\downarrow$ & 5\% ADE$\downarrow$ & & ADE$\downarrow$ & 5\% ADE$\downarrow$ & ADE$\downarrow$ & 5\% ADE$\downarrow$ \\
    \midrule
    Block Stack & \bfseries 0.47 & \bfseries 2.56 & 0.85 & 5.98 & Open Drawer & 2.54 & 16.37 & \bfseries 2.18 & \bfseries 13.19 \\
    Open Microwave & \bfseries 0.68 & \bfseries 6.04 & 1.79 & 6.79 & Right Glass & 3.34 & 30.38 & \bfseries 2.29 & \bfseries 18.55 \\
    Right Glass & 1.96 & 28.82 & \bfseries 1.66 & \bfseries 15.86 & Throw Away Paper & 2.40 & 22.87 & \bfseries 1.47 & \bfseries 8.17 \\
    Pot Lid & 1.18 & 6.89 & \bfseries 0.77 & \bfseries 4.83 & Fold Sock & 1.56 & 10.09 & \bfseries 1.14 & \bfseries 4.68 \\
    \textbf{Average} & \bfseries 1.07 & 11.08 & 1.27 & \bfseries 8.37 & \textbf{Average} & 2.46 & 19.93 & \bfseries 1.77 & \bfseries 11.15 \\
    \bottomrule
    \end{tabular}
\end{table*}

\begin{table*}[t]
\centering
\setlength{\tabcolsep}{2.2pt}
\caption{Simulation success rates (\%) over 200 rollouts per task. Policies
use 20, 50, or 100 action-labeled demonstrations; video-pretrained methods
additionally use 100 actionless videos per task. With 20 demonstrations, \methodname{} achieves the highest success rate
on every task and improves average success by \textbf{29.6} percentage points
over the strongest baseline; it also achieves the highest average success rate at
50 and 100 demonstrations.}
\label{tab:sim_success_rates}
\begin{tabular}{lccccccccccccccc}
\toprule
\textbf{Task} &
\multicolumn{3}{c}{\textbf{AMPLIFY}~\cite{amplify}} &
\multicolumn{3}{c}{\textbf{DP}~\cite{chi_diffusion_2024}} &
\multicolumn{3}{c}{\textbf{ATM}~\cite{wen_any-point_2024}} &
\multicolumn{3}{c}{\textbf{DP3}~\cite{ze_3d_2024}} &
\multicolumn{3}{c}{\textbf{\methodname}} \\
\cmidrule(lr){2-4}
\cmidrule(lr){5-7}
\cmidrule(lr){8-10}
\cmidrule(lr){11-13}
\cmidrule(lr){14-16}
\textbf{\# Demonstrations}
& 20 & 50 & 100
& 20 & 50 & 100
& 20 & 50 & 100
& 20 & 50 & 100
& 20 & 50 & 100 \\
\midrule
Block Stack & \phantom00.5 & \phantom03.5 & \phantom02.0 & 17.0 & 30.5 & 60.5 & 11.0 & \underline{47.5} & \underline{67.0} & \underline{25.5} & 41.5 & 57.0 & \textbf{66.0} & \textbf{84.5} & \textbf{94.0} \\

Right Glass & \phantom02.0 & \phantom06.0 & \phantom09.0 & 16.5 & 27.5 & \underline{76.5} & \phantom08.0 & 25.0 & 32.5 & \underline{21.0} & \underline{43.0} & 70.0 & \textbf{51.0} & \textbf{77.0} & \textbf{97.0} \\

Pot Lid & \phantom03.0 & \phantom04.0 & \phantom07.0 & 42.5 & 51.0 & 58.0 & 27.5 & 51.0 & 54.5 & \underline{51.5} & \underline{60.0} & \underline{77.5} & \textbf{67.5} & \textbf{91.0} & \textbf{93.0} \\

Open Microwave & \phantom07.0 & \phantom02.5 & \phantom08.0 & 17.5 & 35.0 & 81.5 & \underline{70.0} & \textbf{93.5} & \textbf{99.0} & 43.0 & 48.0 & 56.0 & \textbf{75.0} & \underline{78.0} & \underline{89.0} \\

\textbf{Average} & \phantom03.1 & \phantom04.0 & \phantom06.5 & 23.4 & 36.0 & \underline{69.1} & 29.1 & \underline{54.3} & 63.3 & \underline{35.3} & 48.1 & 65.1 & \textbf{64.9} & \textbf{82.6} & \textbf{93.3} \\
\bottomrule
\end{tabular}
\end{table*}

We evaluate \methodname{} in simulation and on a real robot, for both 3D point-track prediction accuracy and
track-conditioned policy learning success rates.

\paragraph{Simulation Setup}
In simulation, we collect demonstrations using a UFactory xArm 7 with a
UFactory parallel-jaw gripper.
We procedurally generate demonstrations by executing end-effector waypoints
derived heuristically from randomly generated initial configurations.
We evaluate on four simulation tasks (Figure~\ref{fig:tasks}),
varying the position and orientation of objects across trials.

\paragraph{Real-World Setup}
For real-world robot demonstrations and evaluation, we use an Xbox controller to
teleoperate the same UFactory xArm 7 and parallel-jaw gripper as in simulation.
For human video demonstration collection, a human performs the manipulation
tasks.
We calibrate a ZED Mini camera to the scene using eye-to-hand calibration, and
keep the camera static.

\begin{figure}[t]
  \centering
  \includegraphics[width=\columnwidth]{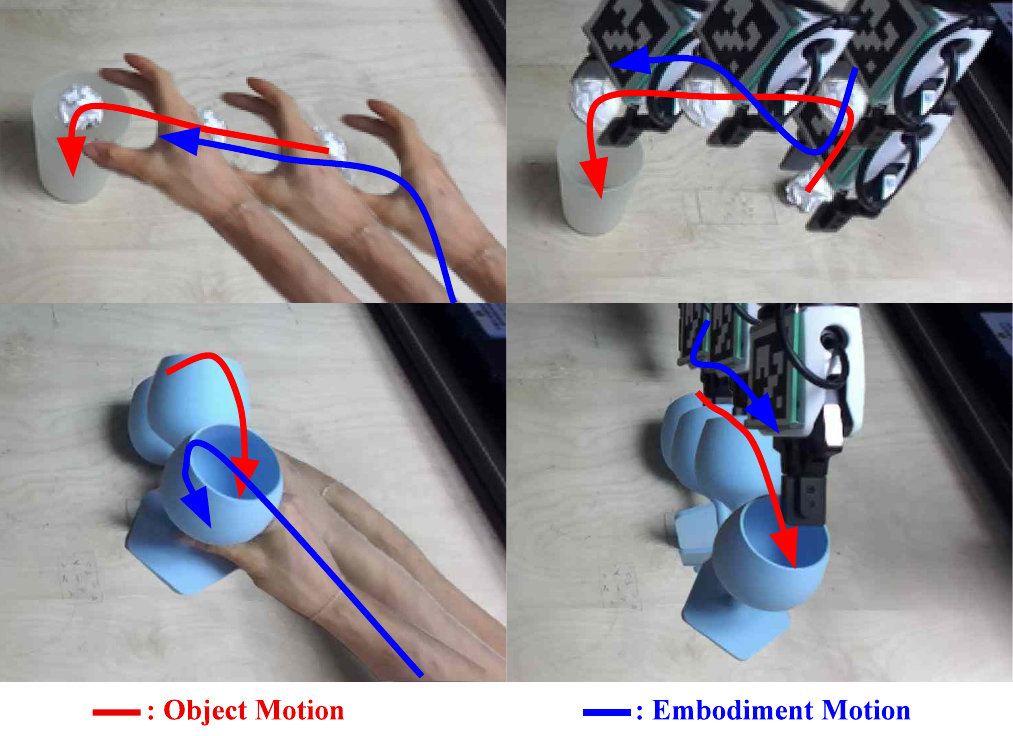}
  \caption{
  Example human vs.\ robot trajectories: \methodname{} learns
  scene motion from human videos (left) and derives a global point-track
  plan, while robot demonstrations teach the closed-loop policy how to realize
  that plan through embodiment-specific actions (right). This separation lets
  the human and robot use different grasps, trajectories, and even strategies.
  \vspace{-10pt}
  }
  \label{fig:trajs}
\end{figure}

We evaluate on four real-world tasks (Figure~\ref{fig:tasks}),
varying the position and orientation of objects across trials.
Although human and robot demonstrations share
similar initial and goal states, the robot gripper and human hand do not need to
have the same configurations or trajectories throughout the task.
Instead of trying to mimic the intended downstream robot motion, the human
demonstrator performs the tasks in a manner they see fit.
For example, for the \emph{Right Glass} task, the human grasps the glass by
the stem, which leads to singular configurations if the robot attempts to
mimic; instead, the robot inserts its gripper into the glass and lifts the rim
such that the glass pivots about its base.
Although the general strategies for the other tasks are more similar between
human and robot, the angles of approach, grasp points, and trajectories 
differ between the human and robot embodiments.
Figure~\ref{fig:trajs} shows examples of these embodiment-induced trajectory differences.

\subsection{Implementation Details}

\paragraph{Data collection}

For both simulation and real-world domains, we train on a mix of action-labeled robot trajectories and actionless videos.
In simulation, we collect actionless videos by removing action labels from robot demonstrations, whereas
in real-world experiments, we collect human demonstration videos.

Our data collection process extracts 3D point tracks for a dense grid of points initialized
at the first image frame of each demonstration. 
The real-world data additionally includes per-frame visibility masks.
In simulation, we obtain tracks from ground-truth rigid-body poses. 
For
real-world videos, we estimate 2D tracks and visibility with
CoTracker3~\cite{karaev_cotracker3_2024}, estimate stereo depth with
FoundationStereo~\cite{wen_foundationstereo_2025}, and lift tracks to 3D.
We segment human-arm points with SAM3~\cite{carion2026sam} and mark them
invisible. We resample all trajectories uniformly in normalized time.
We use coarse trajectories of $\sim$10--16 timesteps as the training targets for the
3D point-track predictor, spanning the complete task execution. We use finer trajectories of
$\sim$30--50 timesteps as the
action targets for the closed-loop policy.
In real-world experiments, we train point-track prediction networks solely on human videos, as we find that 
this provides consistent conditioning for the policies.

\paragraph{Point-track prediction network}
We use a 3-layer MLP with output dimension 256 and GELU activations to tokenize
input points.
We use two decoder blocks with 4 attention heads with head dimension 64.
In between the blocks we use a feed-forward network with expansion ratio 4 and
GELU activation.
After the decoder, a linear head predicts per-point 3D trajectories over the
full task-execution horizon.
At both training and test times, we randomly subsample 2048 points from
the initial point cloud.
We add Gaussian noise with $\sigma=0.01$ to input point clouds as data
augmentation.

\paragraph{Track-conditioned policy network}

The 2048 per-point trajectory tokens are projected to 128-D, and 4 learned
Perceiver-IO queries cross-attend to them to produce 4 point-track tokens. 
These point-track tokens are then used for both global conditioning and
residual cross-attention in the U-Net. 
The global conditioning vector is formed by concatenating three features:
(i) the flattened point-track tokens,
(ii) a 64-D PointNet feature of the current point cloud, and
(iii) a 32-D feature of the current robot configuration,
followed by a projection to 128-D.

The flow-matching action head predicts an action chunk
$A \in \mathbb{R}^{H \times D_A}$ with $D_A = 10$, where $H$ is the prediction horizon.
Each action contains a 3D end-effector position, a 6D continuous orientation
representation~\cite{zhou_continuity_2020}, and a 1D gripper command.
Point clouds and positional actions are jointly centered by the initial point
cloud mean and scaled by its 90th-percentile point radius; rotations are left
unchanged and gripper values use a fixed scale. For multitask learning, we
sample tasks uniformly and, after one warmup epoch, inversely weight each
task's loss by its mean warmup loss.

\begin{table}[t]
\centering
\setlength{\tabcolsep}{2pt}
\caption{Real-world success rates over 20 rollouts per task, using 20 robot
demonstrations and 50 human videos per task. \methodname{} succeeds on
18.8/20 rollouts on average, outperforming the strongest baseline, DP3, by
\textbf{25.0} percentage points and achieving the highest success rate on
every task.}
\label{tab:real_success_rates}
\begin{tabular*}{\columnwidth}{@{\extracolsep{\fill}}lccccc@{}}
\toprule
\textbf{Method} &
\shortstack{\textbf{Open}\\\textbf{Drawer}} &
\shortstack{\textbf{Right}\\\textbf{Glass}} &
\shortstack{\textbf{Throw Away}\\\textbf{Paper}} &
\shortstack{\textbf{Fold}\\\textbf{Sock}} &
\textbf{Average} \\
\midrule
ATM                              & 6/20  & 3/20  & 0/20  & 7/20  & 4.0/20 \\
DP3                              & 14/20 & 18/20 & 9/20  & 14/20 & 13.8/20 \\
\textbf{\methodname}             & \textbf{20/20} & \textbf{20/20} &
                                   \textbf{18/20} & \textbf{17/20} & \textbf{18.8/20} \\
\bottomrule
\end{tabular*}
\end{table}

\subsection{Baselines}

\paragraph{3D Point-Track Prediction Baseline}

We compare to a strong 3D flow prediction method, General Flow~\cite{yuan2024general}, by retraining the General Flow network from scratch on our dataset.

\paragraph{Policy Learning Baselines}
We compare to two behavior cloning baselines that do not use any video pretraining,
and two motion-prediction-based methods pretrained on actionless videos:

\begin{itemize}
  \item \textit{Diffusion Policy (DP)}~\cite{chi_diffusion_2024}: behavior cloning
    with a diffusion action head conditioned on RGB images and robot poses.
  \item \textit{3D Diffusion Policy (DP3)}~\cite{ze_3d_2024}: Diffusion Policy conditioned on 3D point
    clouds instead of images.
  \item \textit{ATM}~\cite{wen_any-point_2024}: predicts future 2D point tracks
    and uses them for policy learning. Following ATM, we train its track
    predictor on both robot and human videos for real-world tasks.
  \item \textit{AMPLIFY}~\cite{amplify}: learns video dynamics in a compressed
    latent space, then learns inverse dynamics from limited action-labeled
    demonstrations.
\end{itemize}

We do not benchmark methods that use embodiment tracks as robot actions or
geometrically convert scene motion into actions in Table~\ref{tab:related_work_comparison}
because they require embodiment correspondences or predefined contacts;
they also directly execute predicted motion, which can be infeasible or collision-prone when
human and robot motion differ. 
We do not benchmark Im2Flow2Act~\cite{xu_flow_2024} because its policy requires simulated play data collected in constructed environments with manually defined exploration primitives, whereas our setting focuses on learning solely from actionless videos and robot demonstrations.

We evaluate all baselines in simulation. We select the best-performing
baseline from each class (DP3 for behavior cloning, and ATM for
flow-based video pretraining) for evaluation on real-world tasks.

\subsection{Results: 3D Point-Track Prediction}
We evaluate \methodname{}'s 3D point-track predictions on all simulation and
real-world tasks using 3D Average Displacement Error (ADE~\cite{yuan2024general})
in millimeters.
We report total ADE over all points and ADE over the 5\,\% of points that
move the most (5\,\% ADE). We propose the 5\,\% ADE metric to reflect task-critical motions, as total ADE is dominated by static background points.
Results are computed over 100 unseen simulation samples per task and 20 unseen
real-world samples per task. 
Although each method predicts full trajectories for every input point, errors
are computed only for visible ground-truth point-timestep pairs $(i,t)$.

Table~\ref{tab:flow_full} shows \methodname\
outperforming General Flow in both metrics on every real-world task,
with average error reductions of
\textbf{28\,\%} and \textbf{44\,\%}, respectively.
The primary advantage of \methodname\ is that it trains on
data General Flow ignores.
Real-world points are often temporarily occluded; General Flow removes
any trajectory with invisible point-timestep pairs during preprocessing, whereas
\methodname\ retains all trajectories and masks losses for individual
invisible point-timestep pairs.
This provides additional supervision over task-critical object points
that are temporarily occluded during manipulation.
For example, in the \emph{Throw Away Paper} task, every paper point is invisible
by the final frame, so General Flow receives no supervision over paper motion.
As a result, \methodname\ generates more informative point tracks from our
real-world manipulation videos.
In simulation, where all points are visible and supervision is identical, \methodname{}'s
single-transformer decoder remains competitive with General Flow's more complex
PointNeXt-conditioned~\cite{qian_pointnext_2022} conditional VAE.

\subsection{Results: Sample-Efficient Policy Learning}
\label{sec:policy_results}

Next, we evaluate \methodname{}'s 3D-point-track-conditioned policy performance.
We display success rates across simulated and real tasks in
Tables~\ref{tab:sim_success_rates} and \ref{tab:real_success_rates}.
We evaluate multitask policies trained on all real-world tasks with 20 robot demonstrations and 50 actionless human videos per task, and separately on all simulation tasks with 20, 50, and 100 action-labeled demonstrations and 100 actionless videos per task. 
Behavior cloning baselines only use action-labeled data.
\methodname{} achieves the highest success rate on every task at 20 robot
demonstrations, and the highest average success rate across tasks at 50 and
100 demonstrations.

Outperforming the behavior cloning baselines (DP and DP3) suggests that
\methodname{} is able to extract useful task information from 3D point-track predictions,
thus effectively leveraging additional supervision from human videos.
We attribute the gains over ATM and AMPLIFY to our design choices,
which we isolate in Section~\ref{sec:results_ablations}.
Overall, these choices yield substantially stronger average performance
while making fewer assumptions about visual and kinematic alignment between
human and robot demonstrations.

\subsection{Results: Ablations}
\label{sec:results_ablations}
We ablate key design decisions across simulated and real tasks.
\begin{table}[t]
\centering
\caption{Ablation of \methodname{} design choices defined in
Section~\ref{sec:results_ablations}.
Simulation variants (a) replace 3D tracks with 2D tracks, replace Perceiver-IO
with mean pooling, remove residual U-Net cross-attention, or train the track
predictor without extra actionless videos. These are trained on 20 robot demonstrations and 100 actionless videos per task, and evaluated on 200 rollouts per task.
Real-world variants (b) include embodiment points in the track targets or
re-predict tracks at every policy query. These are trained on 20 robot demonstrations and 50 human videos per task, and evaluated on 20 rollouts per task.
\methodname{} achieves the highest average success rate in both domains,
demonstrating the contribution of each design choice.
}
\label{tab:ablations}
\setlength{\tabcolsep}{2pt}
\begin{tabular*}{\columnwidth}{@{\extracolsep{\fill}}lccccc@{}}
\toprule
\multicolumn{6}{l}{\textbf{(a) Simulation}} \\
\textbf{Task} &
\shortstack{\textbf{2D}\\\textbf{Tracks}} &
\shortstack{\textbf{No}\\\textbf{Perc.}} &
\shortstack{\textbf{No}\\\textbf{Xattn}} &
\shortstack{\textbf{No}\\\textbf{Videos}} &
\textbf{\methodname} \\
\midrule
Block Stack & 35.0 & 24.5 & 56.0 & \textbf{69.0} & \underline{66.0} \\
Right Glass & 26.5 & 25.0 & \underline{41.0} & 19.5 & \textbf{51.0} \\
Pot Lid & 44.0 & 55.5 & \underline{62.5} & 57.5 & \textbf{67.5} \\
Open Microwave & 37.5 & 29.0 & \underline{70.0} & 60.0 & \textbf{75.0} \\
\textbf{Average} & 35.8 & 33.5 & \underline{57.4} & 51.5 & \textbf{64.9} \\
\bottomrule
\end{tabular*}

\vspace{3pt}
\begin{tabular*}{\columnwidth}{@{\extracolsep{\fill}}lccc@{}}
\toprule
\multicolumn{4}{l}{\textbf{(b) Real world}} \\
\textbf{Task} &
\shortstack{\textbf{Incl. Emb.}\\\textbf{Tracks}} &
\shortstack{\textbf{Per-Step}\\\textbf{Tracks}} &
\textbf{\methodname} \\
\midrule
Open Drawer & 19/20 & 16/20 & \textbf{20/20} \\
Right Glass & \textbf{20/20} & 13/20 & \textbf{20/20} \\
Throw Away Paper & 10/20 & 6/20 & \textbf{18/20} \\
Fold Sock & 12/20 & 13/20 & \textbf{17/20} \\
\textbf{Average} & 15.3/20 & 12.0/20 & \textbf{18.8/20} \\
\bottomrule
\end{tabular*}
\end{table}

\paragraph{Simulation ablations}
In simulation, \methodname{} achieves the highest average success rate across
ablations, as shown in Table~\ref{tab:ablations}.

\begin{itemize}
  \item \textit{No Perceiver-IO}: replaces the Perceiver-IO encoder with mean
    pooling. This produces the largest drop in average success among simulation ablations.
  \item \textit{No U-Net Xattn}: removes residual point-track-token
    cross-attention from the U-Net and reduces performance across all tasks.
  \item \textit{2D Point Tracks}: predicts and conditions on 2D tracks instead of 3D tracks.
  This substantially lowers average success, demonstrating the importance of 3D scene representations.
  \item \textit{No Extra Videos}: trains the track predictor only on the 20
    robot demonstrations. Its strong performance shows that point-track
    prediction itself is a useful inductive bias, while the full model's higher
    average success establishes the benefit of training on extra actionless videos.
\end{itemize}

\paragraph{Real-world ablations}
We conduct two real-world ablations targeting design choices that are specifically relevant when
learning from human videos alongside robot demonstrations. 
For both, we train the track predictor on human and robot videos to reduce train-to-test
distribution shift.
These variants adopt common choices from prior methods
(Table~\ref{tab:related_work_comparison})---using demonstrator
embodiment tracks~\cite{haldar2025point,ren_motion_2025,bharadhwaj_track2act_2024,
lee_tracegen_2026,wen_any-point_2024,amplify}
and re-predicting tracks during execution~\cite{haldar2025point,ren_motion_2025,
yuan2024general,yin_3dmf_2025,wen_any-point_2024,amplify}---to
highlight the benefits of our design.

\begin{itemize}
  \item \textit{Incl.\ Emb.\ Tracks}: includes demonstrator embodiment points
    in the track-prediction targets. Its lower performance is consistent with
    embodiment-specific tracks being less transferable when human and robot
    motion differ.
  \item \textit{Per-Step Tracks}: retains embodiment masking but re-predicts
    tracks from the current observation at every policy query over the action
    horizon $H$, rather than predicting full-task tracks once from the initial
    observation. 
    We attribute its lower performance to two factors: repeated track prediction
    exposing the track predictor to intermediate robot states that differ from
    the human demonstration distribution, and occlusions removing important object and scene point tracks from the policy conditioning.
\end{itemize}

\section{Conclusion}
\label{sec:conclusion}

We introduced \methodname{}, an effective recipe for learning manipulation
policies from human videos with a small number of teleoperated robot
demonstrations. A predictor produces dense 3D tracks over non-embodiment scene
points once as a global plan, a Perceiver-IO encoder compresses them into plan
tokens, and residual cross-attention conditions a closed-loop policy on those
tokens throughout robot execution.

Across our controlled simulation and real-world tasks, this recipe
outperforms behavior-cloning and video-pretraining baselines.
The 3D point-track predictor also improves over a strong flow baseline
on real-world videos, and ablations isolate the contributions of each design choice.
Together, the results establish 3D point-track plans for closed-loop policies as an
effective approach to learning from human videos, even when human and
robot motion differs.

\section{Limitations}
\label{sec:limitations}

Our experiments are conducted across a relatively small set of tasks in a controlled laboratory setting with a static camera.
We therefore do not demonstrate transfer across
differing tasks, viewpoints, or environments. 
The real-world
pipeline also relies on stereo depth estimation, point tracking, and human-arm
segmentation.
We also omit natural language conditioning.

\section{Acknowledgment}
\label{sec:acknowledgment}

This material is based upon work supported by the National Science Foundation Graduate
Research Fellowship Program under Grant Nos. DGE2140739 and DGE2631988. Any opinions,
findings, and conclusions or recommendations expressed in this material are those of the
author(s) and do not necessarily reflect the views of the National Science Foundation.

\FloatBarrier
\bibliographystyle{IEEEtran}
\bibliography{references}

\end{document}